\documentclass[runningheads]{llncs}
\usepackage{graphicx}
\graphicspath{{figures/}}
\usepackage{amsmath,amssymb} % define this before the line numbering.
\usepackage[usenames]{color}
\usepackage[breaklinks=true,colorlinks,bookmarks=false]{hyperref}
\usepackage[width=122mm,left=12mm,paperwidth=146mm,height=193mm,top=12mm,paperheight=217mm]{geometry}
\usepackage{booktabs}
\usepackage{makecell}
\usepackage{wrapfig}
\usepackage{graphicx}
\usepackage{hyperref}
\usepackage{multirow}
\usepackage{makecell}
\usepackage{hyperref}
\usepackage{float}
\usepackage{tabu}

\begin{document}
\title{Analyzing Perception-Distortion Tradeoff using Enhanced Perceptual Super-resolution Network} 

\titlerunning{Enhanced Perceptual Super-resolution Network}

\author{Subeesh Vasu$^1$ \and Nimisha Thekke Madam$^2$ \and Rajagopalan A.N.$^3$}

\authorrunning{Subeesh Vasu, Nimisha T. M., Rajagopalan A.N.}

\institute{Indian Institute of Technology, Madras, India\\
\email{subeeshvasu@gmail.com$^1$,nimiviswants@gmail.com$^2$,raju@ee.iitm.ac.in$^3$}}

\maketitle
\setcounter{footnote}{0}

\begin{abstract}
Convolutional neural network (CNN) based methods have recently achieved great success for image super-resolution (SR). However, most deep CNN based SR models attempt to improve distortion measures (e.g. PSNR, SSIM, IFC, VIF) while resulting in poor quantified perceptual quality (e.g. human opinion score, no-reference quality measures such as NIQE). Few works have attempted to improve the perceptual quality at the cost of performance reduction in distortion measures. A very recent study has revealed that distortion and perceptual quality are at odds with each other and there is always a trade-off between the two. Often the restoration algorithms that are superior in terms of perceptual quality, are inferior in terms of distortion measures. Our work attempts to analyze the trade-off between distortion and perceptual quality for the problem of single image SR. To this end, we use the well-known SR architecture- enhanced deep super-resolution (EDSR) network and show that it can be adapted to achieve better perceptual quality for a specific range of the distortion measure. While the original network of EDSR was trained to minimize the error defined based on per-pixel accuracy alone, we train our network using a generative adversarial network framework with EDSR as the generator module. Our proposed network, called enhanced perceptual super-resolution network (EPSR), is trained with a combination of mean squared error loss, perceptual loss, and adversarial loss. Our experiments reveal that EPSR achieves the state-of-the-art trade-off between distortion and perceptual quality while the existing methods perform well in either of these measures alone.
\end{abstract}
\section{Introduction}
The problem of single image super-resolution (SISR) has attracted much attention and progress in recent years. The primary objective of SISR algorithms is to recover the high-resolution (HR) image from a given single low-resolution (LR) image. By definition, SISR is an ill-posed problem as no unique solution exists for a given LR image. The same LR image can be obtained by down-sampling a large number of different HR images. The ill-posedness of SISR becomes particularly pronounced when the scaling factor increases. Deep learning approaches attempt to solve this ill-posed problem by learning a mapping between the LR and its corresponding HR image in a direct or indirect manner. Recent works on deep neural networks based SISR have shown significant performance improvement in terms of peak signal-to-noise ratio (PSNR).

SISR with deep networks gained momentum with the primal work of Chao et al. \cite{dong2016image}. While \cite{dong2016image} used a 3 layer convolutional neural network (CNN), the subsequent works used deeper network architectures \cite{kim2016accurate,kim2016deeply} and new techniques to improve the restoration accuracy \cite{lim2017enhanced,haris2018deep} and computational complexity \cite{shi2016real,dong2016accelerating}. Despite significant progress in both reconstruction accuracy and speed, a majority of the existing works are still far away from reconstructing realistic textures. This is mainly because of the fact that these works are aimed at improving distortion scores such as PSNR and structural similarity index (SSIM) by optimizing pixel-wise computed error measures such as mean squared error (MSE). In the context of SISR, the optimal MSE estimator returns the mean of many possible solutions \cite{ledig2017photo,sajjadi2017enhancenet} which often leads to blurry, overly smooth, and unnatural appearance in the output, especially at the information-rich regions.  

Previous studies \cite{wang2004image,laparra2016perceptual} revealed that pixel-wise computed error measures correlate poorly with human perception of image quality. Considering the fact that, the behavior of optimization-based SR methods are strongly influenced by the choice of objective function, one should be able to obtain high-quality images by picking the best suited objective function for the task at hand. This is the main motivation behind the recent works on SISR \cite{johnson2016perceptual,ledig2017photo,sajjadi2017enhancenet,mechrez2018learning} that came up with new ways to improve the perceptual quality of reconstructed images. 

A detailed analysis conducted by \cite{Blau_2018_CVPR}  showed that distortion and perceptual quality are at odds with each other and there is always a trade-off between the two. As observed in \cite{Blau_2018_CVPR}, the restoration algorithms that are superior in terms of perceptual quality, are often inferior in terms of distortion measures. They came up with a new methodology for evaluating image restoration methods which can be used to better reveal this trade-off. They have proposed to map SR methods onto a perception-distortion plane and choose the SR method which yields the lowest perceptual score for a given range of distortion measure as the best performing method for that range. They have also suggested that adversarial loss can be used to achieve the desired trade-off for the specific application in mind. Though the work in \cite{Blau_2018_CVPR}  concluded that the existing SISR works perform well in either of these metrics, the possibility to achieve better trade-off in different regions of the perception-distortion plane was left unexplored. 

In this work, we analyze the perception-distortion trade-off that can be achieved by the well-known SISR architecture- enhanced deep super-resolution (EDSR) network \cite{lim2017enhanced}. In our analysis, we limit our focus to SISR by a factor of 4 for LR images distorted by the bicubic down-sampling operator. Selection of EDSR was motivated by the fact that it is one of the state-of-the-art network architecture in terms of the distortion measure for SISR. Since the original work of EDSR proposed in \cite{lim2017enhanced} is aimed at improving distortion measure alone, the perceptual quality achieved by EDSR is poor as pointed out by \cite{Blau_2018_CVPR}. We train EDSR network using a combination of loss functions that can improve distortion measures as well as perceptual quality. Motivated by the observations in \cite{johnson2016perceptual,ledig2017photo,sajjadi2017enhancenet,Blau_2018_CVPR}, we use a combination of MSE loss, perceptual (VGG)  loss, and adversarial loss to train EDSR. Use of adversarial loss to improve perceptual quality allowed our approach to traverse different regions in the perception-distortion plane with ease. We name our approach as enhanced perceptual super-resolution network (EPSR). Our experiments reveal that EPSR can be used to achieve the state-of-the-art trade-off between distortion measure and perceptual quality corresponding to three different regions in the perception-distortion plane.\\
Our main contributions are summarized below.\\
$\bullet$ We expand the scope of EDSR and show that it can be adapted to improve the perceptual quality by compromising on distortion measures.\\
$\bullet$ Our proposed approach achieves the state-of-the-art perception-distortion trade-off results corresponding to different regions in the perception-distortion plane
\section{Related Works}
Though there exist extensive literature studies on multi-image SR \cite{borman1998super,park2003super,farsiu2004fast}, here we limit our discussions to SISR works alone. An overview of recent image SR methods can be found in \cite{nasrollahi2014super,yang2014single}. Early approaches on SISR used sampling theory based interpolation techniques \cite{allebach1996edge,li2001new,zhang2006edge} to recover the lost details. While these algorithms can be very fast, they cannot recover details and realistic textures. Majority of the recent works aim to establish a complex mapping between LR and HR image pairs. The works in \cite{freeman2000learning,freeman2002example} were some of the early approaches to learn such a complex mapping using example-pairs of LR and HR training patches. In \cite{glasner2009super}, the presence of patch redundancies across scales within an image was exploited to generate more realistic textures. This idea was further extended by \cite{huang2015single} wherein self-dictionaries were constructed using self-similar patches that are related through small transformations and shape variations. The convolutional sparse coding framework in \cite{gu2015convolutional} process the whole image and exploits the consistency of neighboring patches to yield better image reconstruction.

To generate edge-preserving realistic textures, \cite{tai2010super} employed a learning-based approach driven by a gradient profile prior. \cite{li2012multi} tried to capture the patch redundancy across different scales using a multi-scale dictionary. HR images from the web with similar contents were used with-in a structure-aware matching criterion to super-resolve landmark images in \cite{yue2013landmark}. The class of neighbor embedding approaches \cite{chang2004super,bevilacqua2012low,gao2012image,timofte2013anchored,timofte2014a+} aim to find similar looking LR training patches from a low dimensional manifold and then combine their corresponding HR patches for resolution enhancement. The overfitting tendency of neighborhood approaches was pointed out by \cite{kim2010single} while also formulating a more generic approach using kernel ridge regression. The work in \cite{dai2015jointly} learned a multitude of patch-specific regressors and proposed to use the most appropriate regressors during testing.
 
Recently, deep neural networks based SR algorithms showed dramatic performance improvements in SISR. Preliminary attempts to deep-learning based SISR appeared in \cite{dong2014learning,dong2016image} 
(SRCNN) wherein a 3 layer network was employed to learn the mapping between the desired HR image and its bicubic up-sampled LR image. This was followed by deeper network architectures \cite{kim2016accurate,kim2016deeply} promising performance improvement over SRCNN. \cite{kim2016accurate} proposed to use residual-learning and gradient clipping with a high-learning rate, whereas \cite{kim2016deeply} relied on a deep recursive layer architecture. The works in \cite{dong2016accelerating,shi2016real} revealed that SR networks can be trained to learn feature representations at the LR dimension itself thereby allowing to use LR images as a direct input rather than using an interpolated image as the input. This improvisation led to significant reduction in computations while maintaining the model capacity and performance gain. To map from the LR feature maps to the final HR image, these works used upsampling modules at the very end of the network. For upsampling, \cite{dong2016accelerating} used a deconv layer whereas \cite{shi2016real}  employed an efficient sub-pixel convolution layer. The work in \cite{ledig2017photo} came up with a deeper architecture made of residual blocks for LR feature learning, called SRResNet. The well-known architecture of EDSR \cite{lim2017enhanced} is built as a modification to SRResNet while using an improvised form of the residual block. They have employed a deeper network architecture with more number of feature units as compared to SRResNet to become the winners of NTIRE2017 \cite{timofte2017ntire}. The work in \cite{haris2018deep}  proposed a deep back-projection network (DBPN) to achieve performance improvement over \cite{timofte2017ntire} for the distortion measure based SISR. It should be noted that all the above-mentioned deep-learning based works have attempted to improve the performance in terms of distortion measures by training loss functions computed in the form of pixel-wise error measures.

Of particular relevance for our paper are the works that have attempted to use loss functions that can better approximate perceptual similarity ensuring recovery of more convincing HR images. The works along this line includes \cite{bruna2015super,johnson2016perceptual,ledig2017photo,sajjadi2017enhancenet,mechrez2018learning,deng2018enhancing}.
Both \cite{bruna2015super} and \cite{johnson2016perceptual} attempted to use an error function derived from the features extracted from a pre-trained VGG network instead of low-level pixel-wise error measures \cite{simonyan2014very}. More specifically, they used the  Euclidean distance between feature maps extracted from the VGG19 network (called VGG loss) as the loss function that was found to give more visually appealing results as opposed to using the MSE loss computed at the pixel-space. SRGAN proposed in \cite{ledig2017photo} was the first attempt to use a GAN-based network which optimizes for the so-called adversarial loss to improve the perceptual quality in SISR. While \cite{ledig2017photo} used a combination of MSE, VGG, and perceptual loss, the work in \cite{sajjadi2017enhancenet} used an additional texture matching loss to generate more realistic textures. \cite{mechrez2018learning} employed contextual loss to replace the perceptual loss for improved perceptual quality. \cite{deng2018enhancing} proposed to combine the high-frequency information of a GAN based method and the content information of an MSE loss based method to obtain achieve the desired balance between distortion and perceptual quality.
\section{Method}
An LR image $I_{\text{LR}}$ can be related to its corresponding HR counterpart ($I_{\text{HR}}$) as
\begin{equation}
I_{\text{LR}}=\text{d}_{\alpha}(I_{\text{HR}})
\label{downsample_eqn}
\end{equation}
where $\text{d}_{\alpha}$ refers to the degradation operator which when acts on $I_{\text{HR}}$ results in $I_{\text{LR}}$ and $\alpha$ ($>$1) is the scaling factor. Though the degrading factors involved in $\text{d}_{\alpha}$ can be a combination of blur, decimation, or noise, in this work, we assume $\text{d}_{\alpha}$ to represent a bicubic downsampling operation with a single scale factor of 4. The task of SISR is to find an approximate inverse $f \approx \text{d}^{-1}$ to yield an HR image estimate $I_{\text{est}}$ from $I_{\text{LR}}$. This problem is highly ill-posed as there exists a large number of possible image estimates $I_{\text{est}}$ for which the degradation relation ($\text{d}_{\alpha}(I_{\text{est}}) = I_{\text{LR}}$) holds true. 

Majority of the deep-learning approaches attempt to find $f$ by minimizing the MSE loss between the network output and the ground truth image ($||I_{\text{est}} - I_{\text{HR}} ||^2_2$). While such a scheme can give excellent results in terms of distortion measures, the resulting images are often blurry and lack high-frequency textures. Previous works on perceptual SR have shown that this limitation can be overcome by employing the loss functions that favor perceptually pleasing results. However, such perceptual improvements result in the reduction of distortion measures. The objective of our work is to experimentally find the perception-distortion trade-off for the state-of-the-art SISR architecture of EDSR.

Next, we will explain the details of our approach, including the network architecture, loss functions, and the methodology that we adopted to find the best possible trade-off corresponding to the network architecture of EDSR.
\begin{figure}
\includegraphics[scale=.4]{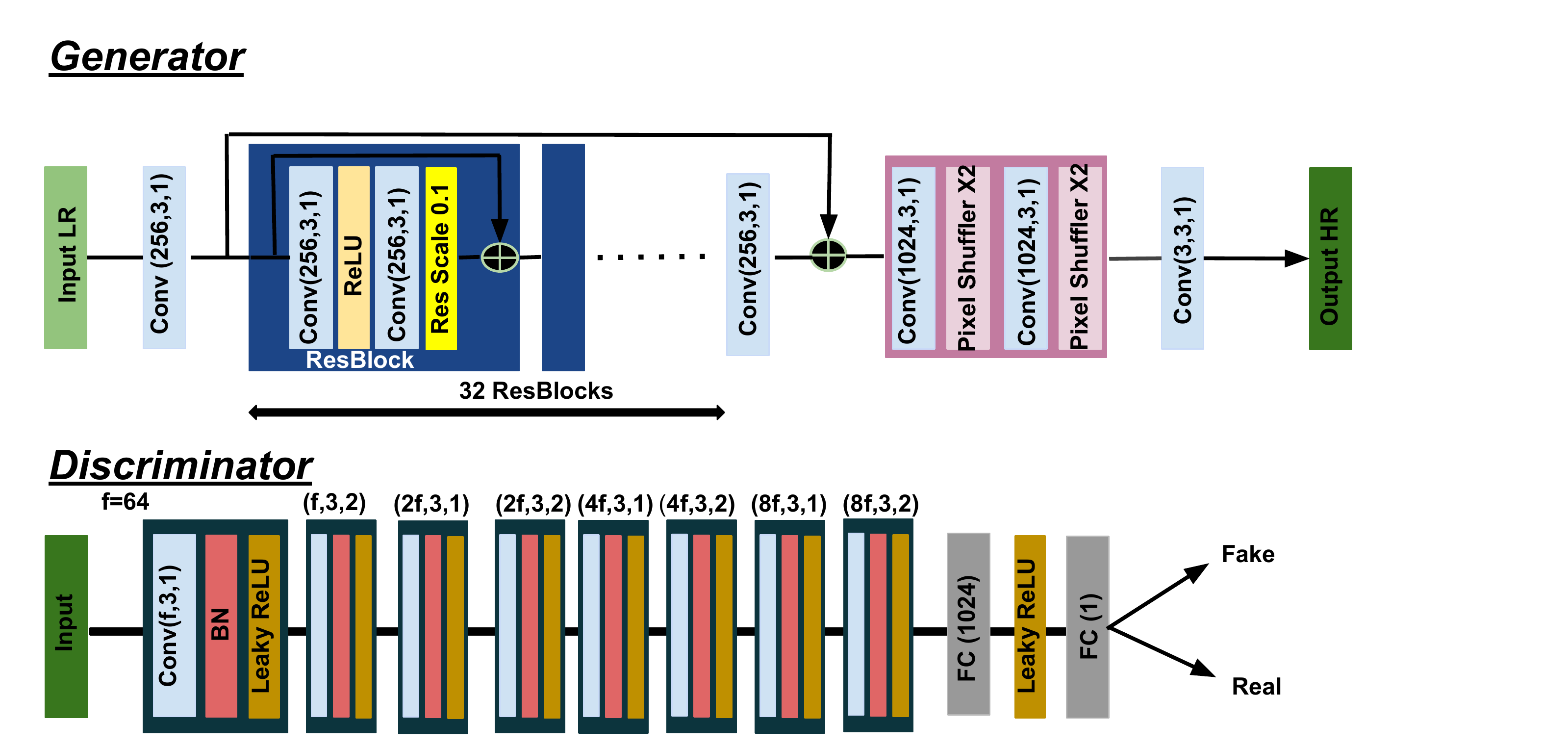}
\caption{Network architecture of EPSR.}
\label{fig1}
\end{figure}
\subsection{Network Architecture} 
The complete architecture of the SR network used in our work is shown in Fig. \ref{fig1}. Our network consists of EDSR acting as the generator module and a CNN based classifier acting as a discriminator module. In the diagram shown in Fig. \ref{fig1}, $\textbf{conv}(\textbf{n},\textbf{k},\textbf{s})$ refers to a convolution layer with $\textbf{n}$ number of $\textbf{k} \times \textbf{k}$ filters performing convolution by a stride factor of $\textbf{s}$. For simplicity we use the notation $(\textbf{n},\textbf{k},\textbf{s})$ instead of $\textbf{conv}(\textbf{n},\textbf{k},\textbf{s})$ in the diagram corresponding to the discriminator. EDSR is built based on a modified form of residual block wherein no batch normalization units are used. An additional residual scaling layer (multiplication by a constant scale factor of 0.1) is inserted onto each residual block to numerically stabilize the training procedure in the absence of batch-normalization. This kind of a modified form of the residual block has allowed the work in \cite{lim2017enhanced} to employ a deeper network architecture with more number of feature units in each layer to improve the performance over the SRResNet architecture of \cite{ledig2017photo}. The generator module comprises of 32 (modified form of) residual blocks (refer Fig. \ref{fig1} for more details). The LR images are directly provided to the network as inputs. To increase the resolution by a factor of 4, residual blocks are followed by two pixel shuffler units each of which increases the spatial resolution by a factor of 2. As shown in Fig. \ref{fig1}, the discriminator that we used is a 10 layer network trained to produce a single output 0/1 depending on the input data which can be $I_{\text{est}}/I_{\text{HR}}$. The network consists of a number of convolution layers followed by fully connected layers which map from an image to a single output value.
\subsection{Training and Loss Functions} 
We used the following form of loss function to train the network.
\begin{equation}
\mathcal{L}= \lambda_1 \mathcal{L_{VGG}} + \lambda_2 \mathcal{L}_{E} + \lambda_3 \mathcal{L}_{adv}
\end{equation}
where $\mathcal{L}$ is the total loss function used for training the generator network. $\mathcal{L}_{E}$ is the MSE between the network output and ground truth HR image given by
\begin{equation}
\mathcal{L}_{E}= ||I_{\text{est}}-I_{\text{HR}}||^2_2
\label{mse_loss}
\end{equation}
$\mathcal{L_{VGG}}$ is the perceptual loss \cite{johnson2016perceptual} computed using the VGG network \cite{simonyan2014very} as
\begin{equation}
\mathcal{L_{VGG}}= ||\phi (I_{\text{est}})-\phi (I_{\text{HR}})||^2_2
\label{vgg_loss}
\end{equation}
where $\phi$ refers to VGG feature layers. Previous studies on perceptual SISR \cite{ledig2017photo,sajjadi2017enhancenet} have shown that the use of perceptual loss $\mathcal{L_{VGG}}$ can provide further boost in the detail enhancement if used along with adverserial loss. Following this line, we also use $\mathcal{L_{VGG}}$ to provide an additional support to the adversarial loss for improving the perceptual quality. Similar to the work in \cite{ledig2017photo}, we used VGG54 as the feature extraction layer (i.e, the feature maps obtained by the 4th convolution (after activation) and before the 5th max-pooling layer). $\mathcal{L}_{adv}$ is the adversarial loss derived from the discriminator network and is given by
\begin{equation}
\mathcal{L}_{adv}= - \log D(G(I_{\text{LR}}))
\label{adv_loss}
\end{equation}
where $G(\cdot)$ and $D(\cdot)$ indicates the network outputs from the generator and discriminator respectively. $\lambda_1$, $\lambda_2$, and $\lambda_3$ are non-negative scale factors that can be varied to control the perception-distortion trade-off.

Motivated from the observation that GANs can provide a principled way to approach the perception-distortion bound \cite{Blau_2018_CVPR}, we train our network using different values of $\lambda_2$ and $\lambda_3$ (refer Table \ref{region-table}) to achieve the best possible perception-distortion trade-off using EPSR. The training of EPSR is done similar to that of \cite{ledig2017photo}. The generator network is trained to learn a mapping from input image $I_{\text{LR}}$ to an approximate estimate of the HR image $I_{\text{est}}$ by optimizing the loss function $\mathcal{L}$. Simultaneously, the discriminative network $D$ is trained to distinguish between real images $I_{\text{HR}}$ from the training dataset and generated image estimates of the network $G(I_{\text{LR}})$. To train the discriminator we minimize the loss function.
\begin{equation}
\mathcal{L}_D = - \text{log}(D(I_\text{HR})) - \text{log}(1 - D(G(I_\text{LR})))
\label{d_loss}
\end{equation}
During training, the discriminator was updated twice followed by a single generator update. Also, to train the network with different values of $\lambda_2$ and $\lambda_3$, we initialized the model weights of generator using pre-trained weights of EDSR (obtained by training EDSR with $\lambda_1 = \lambda_3 = 0$).
\begin{table}[t]
\centering
\begin{tabular}{|c|c|c|c|c|c|c|} \hline
\textbf{Network model $\rightarrow$} & \multicolumn{3}{c|}{BNet} & \multicolumn{3}{c|}{EPSR}\\
\hline
\textbf{}& $\lambda_1$&$\lambda_2$&$\lambda_3$&$\lambda_1$&$\lambda_2$&$\lambda_3$\\ \hline
Region 1 (RMSE $\leq$ 11.5) &1 &0.1&0.4&1&.05&0.4\\
\hline
Region 2 (11.5 $<$ RMSE $\leq$ 12.5) &1&0.05&0.4&1&0.02&0.4\\
\hline
Region 3 (12.5 $<$ RMSE $\leq$ 16) &1&0.0005&0.6&1&0.0005&0.6\\
\hline
\end{tabular}
\caption{Parameter settings used for training BNet and EPSR to obtain results corresponding to Region 1, 2, and 3. BNet (refer Section \ref{sectn_evaln}) is a baseline network used for performance comparison.}
\label{region-table}
\end{table}
\subsection{Implementation Details}
To train our network, we used the first 800 images of DIV2K dataset \cite{agustsson2017ntire}. The HR images were bicubically down-sampled by a factor of 4 to create the input LR images for training. We followed a patch-wise training wherein the patch-size of the network output was set to 192. We used ADAM \cite{kingma2014adam} optimizer with a momentum of 0.9 and a batch size of 4. The network was trained for 300 epochs and the learning rate was initially set to 5e-5 which was reduced by a factor of 0.5 after 150 epochs. We used pre-trained VGGNet weights to enforce the effect of perceptual loss. Our implementation was done in PyTorch and was built on top of the official PyTorch implementation of \cite{lim2017enhanced} which was available online. The code was run on TITAN-X Pascal GPU. It took around 45 hrs to complete the training of one single network. On an average, during testing, to super-resolve an input image of size 100 $\times$ 100, EPSR takes around 0.5 seconds.
\section{Evaluation}
\label{sectn_evaln}
To evaluate the performance, we follow a procedure similar to that of ``The PIRM challenge on perceptual super-resolution'' (PIRM-SR) \cite{2018arXiv180907517B} and \cite{Blau_2018_CVPR}. The evaluation is done in a perceptual-quality aware manner \cite{Blau_2018_CVPR}, and not based solely on the basis of distortion measures. To this end, we divide the perception-distortion plane \cite{Blau_2018_CVPR} into three regions defined by thresholds on the RMSE of the SR outputs. The thresholds used for the three regions are mentioned in Table \ref{region-table}.

\begin{table}[t]
\centering
\begin{tabular}{|c|c|c|c|c|c|c|c|c|c|} \Xhline{1.5pt}
\textbf{Dataset}& Scores &bicubic &SRCNN\cite{dong2016image} &EDSR\cite{lim2017enhanced} &DBPN\cite{haris2018deep} & BNet$_1$ & EPSR$_1$\\ 
\Xhline{1.5pt}
\multirow{4}{*}{\textbf{PIRM-self}}
&RMSE & 13.2923& 12.0194& 10.8934& 10.9779&11.4956&11.4924\\\cline{2-8}
 & PSNR& 26.5006& 27.5258& 28.5754& 28.4927&27.9752&27.9852\\ \cline{2-8}
&SSIM & 0.6980& 0.7429& 0.7808& 0.7773&0.7511&0.7508\\ \cline{2-8}
&PI & 6.805& 5.8247& 5.0399& 5.2043&\textcolor{red}{4.1492}&\textcolor{red}{\textbf{2.9459}}\\
\Xhline{1.5pt}
\multirow{3}{*}{\textbf{Set5}} &PSNR & 28.4164& 30.5314& 32.4034& 32.3337&31.4505&31.6954\\ \cline{2-8}
&SSIM & 0.8096& 0.8630& 0.8960& 0.8949&0.8739&0.8751\\ \cline{2-8}
&PI & 7.323& 7.0858& 5.8366 & 6.107 &\textcolor{red}{5.4136}&\textcolor{red}{\textbf{4.8087}}\\ 
\Xhline{1.5pt}
\multirow{3}{*}{\textbf{Set14}} &PSNR & 25.6675& 26.7191& 27.4193& 28.1266&27.0541&27.0123\\  \cline{2-8}
&SSIM & 0.6921& 0.7316& 0.7543& 0.7686&0.7342&0.7315\\  \cline{2-8}
&PI & 6.968& 6.0189& 5.2942 & 5.5723 &\textcolor{red}{4.4824}&\textcolor{red}{\textbf{3.7101}}\\ 
\Xhline{1.5pt}
\multirow{3}{*}{\textbf{BSD100}} & PSNR & 26.2128& 26.7564& 27.0088& 27.0145&26.8711&26.7497\\  \cline{2-8}
&SSIM & 0.6839& 0.7198& 0.7396& 0.7364&0.71782&0.7133\\  \cline{2-8}
&PI & 6.9485&5.9707 &5.36 &5.5362 &\textcolor{red}{4.6416}&\textcolor{red}{\textbf{3.5503}}\\ 
\Xhline{1.5pt}
\multirow{3}{*}{\textbf{Urban100}} &PSNR & 22.7809& 23.5834& 24.5753& 24.4825&24.1029&24.3012\\  \cline{2-8}
&SSIM & 0.6477& 0.6984& 0.7517& 0.7460&0.72199&0.7302\\  \cline{2-8}
&PI & 6.8796& 5.8414&5.0395 &5.1944 &\textcolor{red}{4.2223}&\textcolor{red}{\textbf{3.8994}}\\
\Xhline{1.5pt}
\end{tabular}
\caption{Results on public benchmark test data and PIRM-self validation data for existing distortion measure specific methods and our methods corresponding to region 1 (BNet$_1$ and EPSR$_1$). Bold red indicates the best performance in Region 1 and light red indicates the second best.}
\label{table1}
\end{table}

We used perceptual index (PI) to quantify the perceptual quality. PI is computed by combining the quality measures of Ma-score \cite{ma2017learning} and NIQE \cite{mittal2013making} as follows
\begin{equation}
\text{PI} = \dfrac{1}{2}((10-\text{Ma-score})+\text{NIQE})
\label{PI}
\end{equation}
Note that, a lower PI indicates better perceptual quality. The algorithm with the best perceptual score (or equivalently lowest PI) in each region is treated as the one with most visually pleasing results corresponding to that particular region. This approach of region-wise comparison quantifies the accuracy and perceptual quality of algorithms jointly, and will, therefore, enable a fair comparison of perceptual-driven methods alongside algorithms that target PSNR maximization.

\begin{table}[t]
\centering
\begin{tabular}{|c|c|c|c|c|c|c|c|c|c|c|}\Xhline{1.5pt}
\textbf{Dataset}& Scores &ENet\cite{sajjadi2017enhancenet} & CX\cite{mechrez2018learning} & BNet$_2$ & BNet$_3$ & EPSR$_2$ & EPSR$_3$\\ 
\Xhline{1.5pt}
\multirow{4}{*}{\textbf{PIRM-self}} &RMSE & 15.9853 & 15.2477 & 12.4709&15.6292& 12.4094& 15.3586\\ \cline{2-8}
&PSNR  & 25.0642 & 25.4051 &27.1789&25.2845& 27.342& 25.4541\\  \cline{2-8}
&SSIM  & 0.6463& 0.6744 & 0.7184&0.6560& 0.72744& 0.6655\\  \cline{2-8}
&PI & 2.6876& \textcolor{red}{2.131} & \textcolor{blue}{2.4795}&2.2354&\textcolor{blue}{\textbf{2.3881}}& \textcolor{red}{\textbf{2.0688}}\\
\Xhline{1.5pt}
\multirow{3}{*}{\textbf{Set5}} &PSNR & 28.5641&29.1017 & 30.7637&28.6764& 31.2168& 29.5757\\  \cline{2-8}
&SSIM  & 0.80819&0.82982 & 0.85485&0.80948& 0.8630&0.8388\\  \cline{2-8}
&PI & \textcolor{red}{\textbf{2.9261}} &3.2947  &\textcolor{blue}{\textbf{4.0003}}&\textcolor{red}{3.2223}&\textcolor{blue}{4.1123}&3.2571\\
\Xhline{1.5pt}
\multirow{3}{*}{\textbf{Set14}} &PSNR &25.7521 & 25.2265 &26.5242&25.2487& 26.6068& 25.5238\\  \cline{2-8}
&SSIM  &0.67953 & 0.67606 &0.7104&0.6595& 0.71342&0.6848\\  \cline{2-8}
&PI & 3.014& 2.759 & \textcolor{blue}{3.1706}&\textcolor{red}{\textbf{2.6473}}& \textcolor{blue}{\textbf{3.0246}}&\textcolor{red}{2.6982}\\
\Xhline{1.5pt}
\multirow{3}{*}{\textbf{BSD100}} &PSNR & 25.3764 & 24.2868 &26.1619&24.7761&26.2819&24.9753\\  \cline{2-8}
&SSIM  & 0.64268 & 0.6396 &0.6826&0.6217& 0.69054&0.64503\\  \cline{2-8}
&PI & 2.9297& \textcolor{red}{2.2501} & \textcolor{blue}{2.801}&2.3674& \textcolor{blue}{\textbf{2.7458}}&\textcolor{red}{\textbf{2.199}}\\
\Xhline{1.5pt}
\multirow{3}{*}{\textbf{Urban100}} &PSNR & 23.6771 & 22.8444 &23.5657&22.0168& 23.9985&22.7959\\  \cline{2-8}
&SSIM  & 0.69775 &0.6748 &0.6934&0.6454& 0.71798 &0.66631\\  \cline{2-8}
&PI & 3.4679& 3.3894 & \textcolor{blue}{3.6345}&\textcolor{red}{\textbf{3.2721}}& \textcolor{blue}{\textbf{3.6236}}&\textcolor{red}{3.3316}\\
\Xhline{1.5pt}
\end{tabular}
\caption{Results on public benchmark test data and PIRM-self for existing perceptual quality specific methods and our proposed methods corresponding to Region 2 and Region 3 (EPSR$_2$ and EPSR$_3$). Bold blue (and red) indicates the best performance in Region 2 (and Region 3) and light blue (and red) indicates the second best.}
\label{table2}
\end{table}
To have an idea about the performance level of EPSR, we compare it with that of the trade-off values achieved by a baseline network formed by our-self. We call our baseline network as BNet and is a simplified form of EPSR. Unlike EPSR, the generator of BNet has no residual scaling. BNet uses 32 number of residual blocks and 64 filters in each layer of the residual block. BNet is equivalent to the network in \cite{ledig2017photo} (SRGAN) except for the fact that \cite{ledig2017photo} use batch normalization units in the generator whereas BNet does not.
%%%%%%%%%%IMAGE1%%%%%%%%%%%%%%%%%%%%%%%%
\begin{figure*}[t]
\centering
\begin{tabular}{cc}
\begin{tabular}{c}
\includegraphics[width=.28\linewidth, height=.28\linewidth]{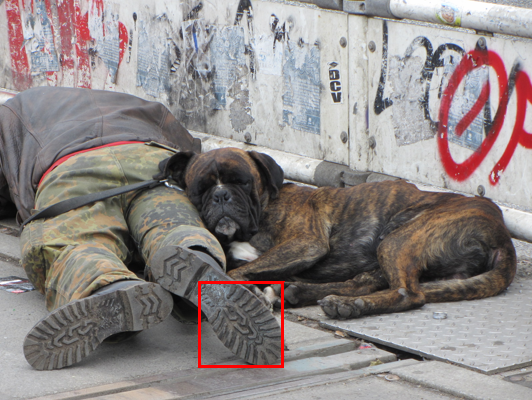}\\
\scriptsize 042 from PIRM-self\\
\end{tabular}&
\begin{tabular}{c}
\begin{tabular}{cccccc}
\includegraphics[width=.12\linewidth]{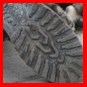}&
\includegraphics[width=.12\linewidth]{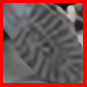}&
\includegraphics[width=.12\linewidth]{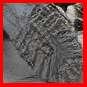}&
\includegraphics[width=.12\linewidth]{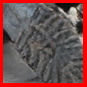}&
\includegraphics[width=.12\linewidth]{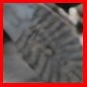}
\\
\scriptsize $I_{\text{HR}}$ &\scriptsize EDSR\cite{lim2017enhanced} & \scriptsize ENet\cite{sajjadi2017enhancenet} & \scriptsize CX\cite{mechrez2018learning} &\scriptsize \scriptsize BNet$_1$\\
\includegraphics[width=.12\linewidth]{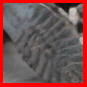}&
\includegraphics[width=.12\linewidth]{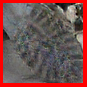}&
\includegraphics[width=.12\linewidth]{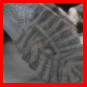}&
\includegraphics[width=.12\linewidth]{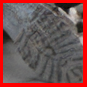}&
\includegraphics[width=.12\linewidth]{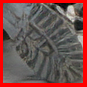}\\
\scriptsize BNet$_2$ & \scriptsize BNet$_3$ & \scriptsize EPSR$_1$ & \scriptsize EPSR$_2$ & \scriptsize EPSR$_3$\\
\end{tabular}\\
\end{tabular}\\
%%%%%%%%%%%% IMAGE2%%%%%%%%%%%%%%%%%%%%%%%%%%%%%%
\begin{tabular}{c}
\includegraphics[width=.28\linewidth, height=.28\linewidth]{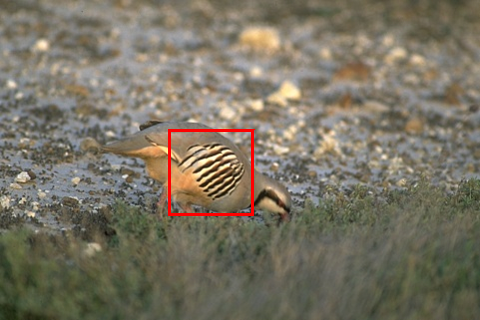}\\
\scriptsize 8023 from BSD100\\
\end{tabular}&
\begin{tabular}{c}
\begin{tabular}{cccccc}
\includegraphics[width=.12\linewidth]{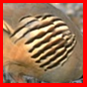}&
\includegraphics[width=.12\linewidth]{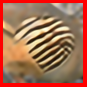}&
\includegraphics[width=.12\linewidth]{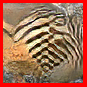}&
\includegraphics[width=.12\linewidth]{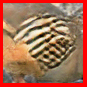}&
\includegraphics[width=.12\linewidth]{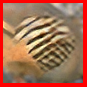}
\\
\scriptsize $I_{\text{HR}}$ &\scriptsize EDSR\cite{lim2017enhanced} & \scriptsize ENet\cite{sajjadi2017enhancenet} & \scriptsize CX\cite{mechrez2018learning} &\scriptsize \scriptsize BNet$_1$\\
\includegraphics[width=.12\linewidth]{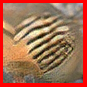}&
\includegraphics[width=.12\linewidth]{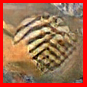}&
\includegraphics[width=.12\linewidth]{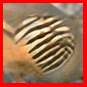}&
\includegraphics[width=.12\linewidth]{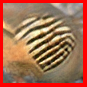}&
\includegraphics[width=.12\linewidth]{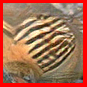}\\
\scriptsize BNet$_2$ & \scriptsize BNet$_3$ & \scriptsize EPSR$_1$ & \scriptsize EPSR$_2$ & \scriptsize EPSR$_3$\\
\end{tabular}\\
\end{tabular}\\
%%%%%%%%%%%% IMAGE3%%%%%%%%%%%%%%%%%%%%%%%%%%%%%%
\begin{tabular}{c}
\includegraphics[width=.28\linewidth, height=.28\linewidth]{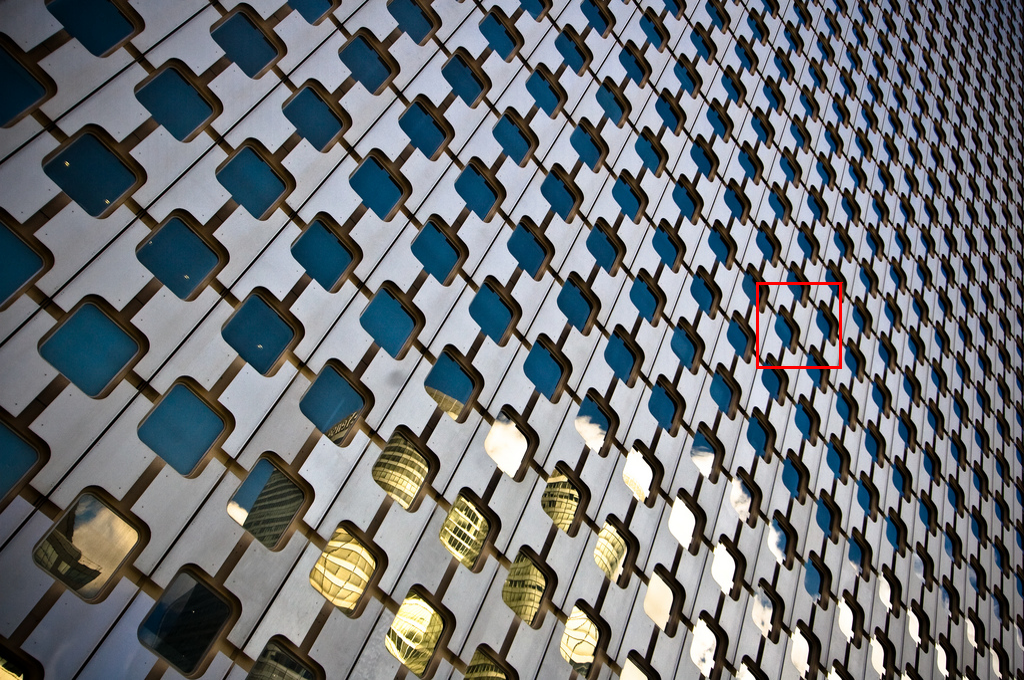}\\
 \scriptsize 041 from Urban 100\\
\end{tabular}& 
\begin{tabular}{c}
\begin{tabular}{cccccc}
\includegraphics[width=.12\linewidth]{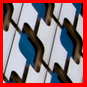}&
\includegraphics[width=.12\linewidth]{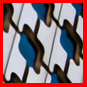}&
\includegraphics[width=.12\linewidth]{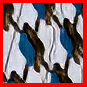}&
\includegraphics[width=.12\linewidth]{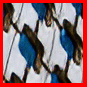}&
\includegraphics[width=.12\linewidth]{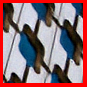}\\
\scriptsize $I_{\text{HR}}$ &\scriptsize EDSR\cite{lim2017enhanced} & \scriptsize ENet\cite{sajjadi2017enhancenet} & \scriptsize CX\cite{mechrez2018learning} &\scriptsize \scriptsize BNet$_1$\\
\includegraphics[width=.12\linewidth]{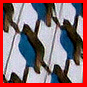}&
\includegraphics[width=.12\linewidth]{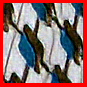}&
\includegraphics[width=.12\linewidth]{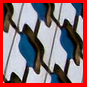}&
\includegraphics[width=.12\linewidth]{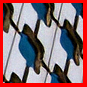}&
\includegraphics[width=.12\linewidth]{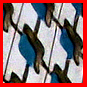}\\
\scriptsize BNet$_2$ & \scriptsize BNet$_3$ & \scriptsize EPSR$_1$ & \scriptsize EPSR$_2$ & \scriptsize EPSR$_3$\\
\end{tabular}\\
\end{tabular}\\
\end{tabular}\\
\caption{Qualitative comparison of our models with other works on x4 super-resolution.
The image examples are taken from datasets of PIRM-self (Row 1), BSD100 (Row 2),
and Urban100 (Row 3).}
\label{qual1}
\end{figure*}

To perform a region-wise comparison, we train both BNet and EPSR with a different set of weights for MSE loss and adversarial loss. The weights for the best trade-off was empirically found for each region (refer to Table \ref{region-table} for details). In the following comparisons, BNet$_1$ (/BNet$_2$/BNet$_3$) and EPSR$_1$ (/EPSR$_2$/EPSR$_3$) refers to the best model weights (i.e., the ones with the lowest PI) obtained for Region 1 (/2/3) corresponding to BNet and EPSR respectively. We perform the region-wise performance comparisons with the most relevant methods on distortion measure (bicubic interpolation, \cite{dong2016image,lim2017enhanced,haris2018deep}) as well as perceptual quality \cite{sajjadi2017enhancenet,mechrez2018learning}. Since the code of SRGAN \cite{ledig2017photo} was not available, an equivalent comparison is done using BNet. We could not compare with the other perceptual SR methods \cite{bruna2015super,johnson2016perceptual,deng2018enhancing}, as the source codes for them were not available.

Evaluation is done on the public benchmark data sets of Set5 \cite{bevilacqua2012low}, Set14 \cite{zeyde2010single}, BSD100 \cite{martin2001database}, Urban100 \cite{huang2015single} and the self-validation data from PIRM-SR (PIRM-self) \cite{2018arXiv180907517B}. Since PIRM-self contains 100 images with an equal distribution of scenes and quality, it can be treated as the most suited dataset for perceptual quality-based evaluation. Consequently, we use the average MSE values computed over PIRM-self to define the three regions in the perception-distortion plane.
\subsection{Quantitative Results}
To quantitatively compare the performance, we report the values of PSNR, SSIM, and PI. The results corresponding to \cite{lim2017enhanced} is obtained using the model weights of EDSR obtained through our own training. Also, the values that we have obtained for the existing methods on distortion measure is slightly different as compared to the ones reported in the original papers. This could be due to the difference in the way we have computed the scores. All the scores reported in this paper are computed on the y-channel after removing a 4-pixel border.

Table \ref{table1} lists the quantitative comparison of distortion measure based methods with that of BNet$_1$ and EPSR$_1$. \footnote{Bicubic and SRCNN correspond to Region 2 since their RMSE values are above 11.5} As is evident from Table \ref{table1}, EPSR performs the best and achieve the lowest PI in Region 1 and BNet turns out to be the second best. Both BNet$_1$ and EPSR$_1$ is able to deliver low PI values (i.e., better perceptual quality) while maintaining much better distortion measures (RMSE, PSNR, and SSIM) as compared to bicubic interpolation and SRCNN. A careful inspection of the distortion measure based method reveals that the perceptual quality improves as the PSNR increases, however, the relative improvement is very narrow. Differently, a comparison between EDSR and EPSR$_1$ shows that the use of adversarial loss has helped EPSR$_1$ to achieve significant improvement in perceptual quality but while subjected to reduction in distortion measures. 

Table \ref{table2} lists the quantitative comparison of perceptual-SISR methods with that of BNet and EPSR corresponding to Region 2 and 3. It should be noted that, among all the datasets that we have compared, Set5, Set14, and Urban 100 are not the ideal ones for perceptual quality comparisons. Because Set5 and Set14 have only a small number of images whereas Urban100 covers only the images of urban scenes. Both, BSD 100 and PIRM-self covers wide-variety of scenes and can be treated as an ideal collection of natural images of different kinds. Comparisons over BSD 100 and PIRM-self in Table \ref{table2} underscore the superior perceptual quality improvement achieved by EPSR. In other datasets, the method which has the lowest PI varies. In Set5, ENet\cite{sajjadi2017enhancenet} performs best in Region 3, whereas BNet$_2$ performs best in Region 2. In Set14 and Urban 100, the best performing methods are CX\cite{mechrez2018learning}, BNet, and EPSR with only a comparable performance difference between each other. 

Considering all regions together, one can see that, EPSR achieves the best perceptual scores, with CX\cite{mechrez2018learning} being second best. By comparing BNet and EPSR scores across different regions we can notice the trade-off between the PI and RMSE. When we allowed having more distortion (i.e., higher RMSE), both BNet and EPSR are able to yield significant improvement in perceptual quality. Note that the generator network of BNet is inferior to that of EPSR in terms of distortion measures. This allows EPSR to achieve better perceptual quality than BNet for a fixed level of distortion. We believe the following as the primary reason for such an effect. To improve the perceptual quality, a network needs to generate more realistic textures resulting in an increase of the content deviation from the ground truth image. Therefore, for a given distortion range, a generator network which is superior in terms of distortion-measure is more likely to generate results with the best perceptual quality when trained using a GAN framework.
%%%%%%%%%%IMAGE1%%%%%%%%%%%%%%%%%%%%%%%%
\begin{figure*}
\centering
\begin{tabular}{cc}
\begin{tabular}{c}
\includegraphics[width=.3\linewidth, height=.3\linewidth]{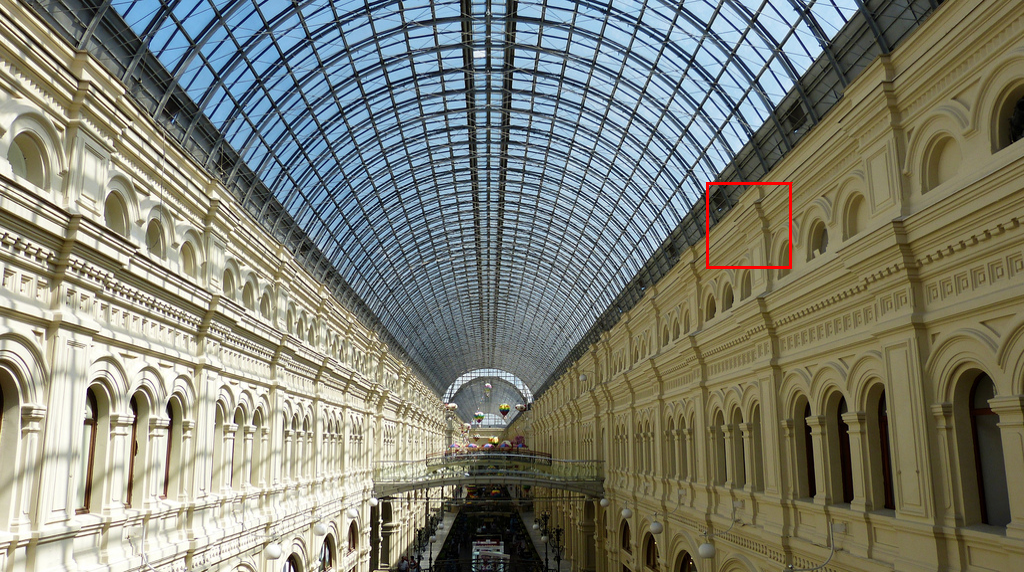}\\
\scriptsize 008 from Urban100\\
\end{tabular}&
\begin{tabular}{c}
\begin{tabular}{cccccc}
\includegraphics[width=.12\linewidth]{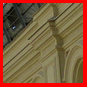}&
\includegraphics[width=.12\linewidth]{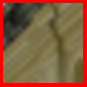}&
\includegraphics[width=.12\linewidth]{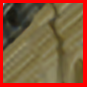}&
\includegraphics[width=.12\linewidth]{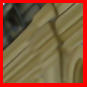}&
\includegraphics[width=.12\linewidth]{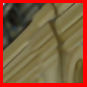}
\\
\scriptsize $I_{\text{HR}}$ &\scriptsize Bicubic & \scriptsize SRCNN\cite{dong2016image} & \scriptsize EDSR\cite{lim2017enhanced} &\scriptsize  DBPN\cite{haris2018deep}\\
\includegraphics[width=.12\linewidth]{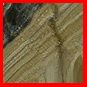}&
\includegraphics[width=.12\linewidth]{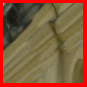}&
\includegraphics[width=.12\linewidth]{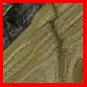}&
\includegraphics[width=.12\linewidth]{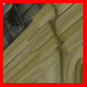}&
\includegraphics[width=.12\linewidth]{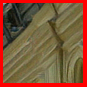}\\
\scriptsize ENet\cite{sajjadi2017enhancenet} & \scriptsize BNet$_1$ & \scriptsize BNet$_3$ & \scriptsize EPSR$_1$ & \scriptsize EPSR$_3$\\
\end{tabular}\\
\end{tabular}\\
%%%%%%%%%%IMAGE2%%%%%%%%%%%%%%%%%%%%%%%%
\begin{tabular}{c}
\includegraphics[width=.3\linewidth, height=.3\linewidth]{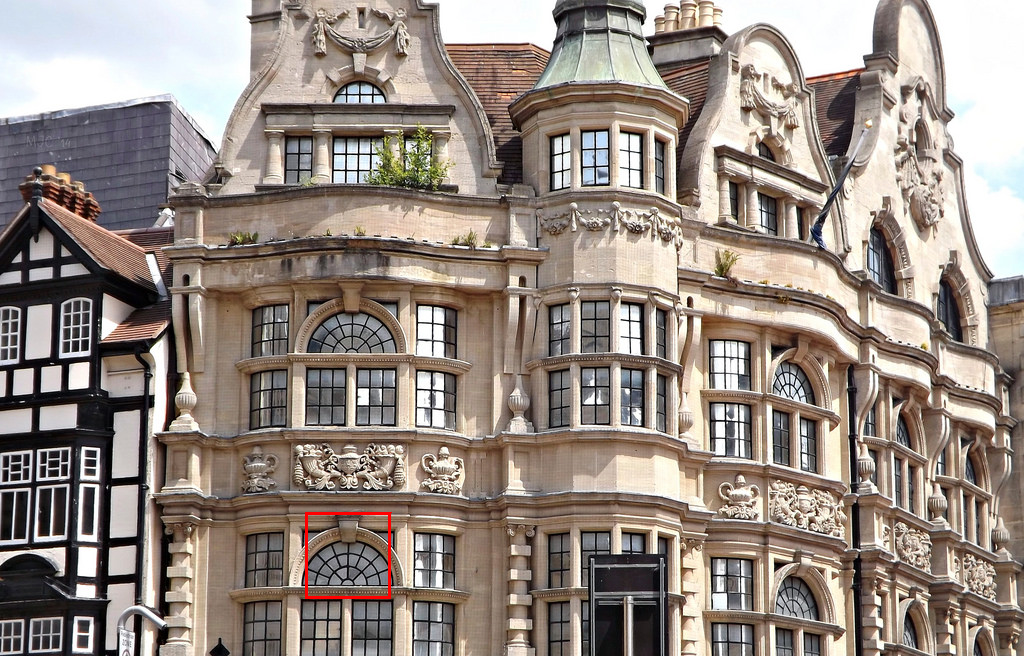}\\
\scriptsize 053 from Urban100\\
\end{tabular}&
\begin{tabular}{c}
\begin{tabular}{cccccc}
\includegraphics[width=.12\linewidth]{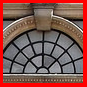}&
\includegraphics[width=.12\linewidth]{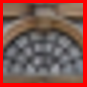}&
\includegraphics[width=.12\linewidth]{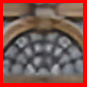}&
\includegraphics[width=.12\linewidth]{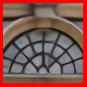}&
\includegraphics[width=.12\linewidth]{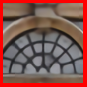}
\\
\scriptsize $I_{\text{HR}}$ &\scriptsize Bicubic & \scriptsize SRCNN\cite{dong2016image} & \scriptsize EDSR\cite{lim2017enhanced} &\scriptsize  DBPN\cite{haris2018deep}\\
\includegraphics[width=.12\linewidth]{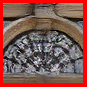}&
\includegraphics[width=.12\linewidth]{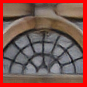}&
\includegraphics[width=.12\linewidth]{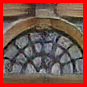}&
\includegraphics[width=.12\linewidth]{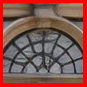}&
\includegraphics[width=.12\linewidth]{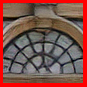}\\
\scriptsize ENet\cite{sajjadi2017enhancenet} & \scriptsize BNet$_1$ & \scriptsize BNet$_3$ & \scriptsize EPSR$_1$ & \scriptsize EPSR$_3$\\
\end{tabular}\\
\end{tabular}\\
%%%%%%%%%%IMAGE3%%%%%%%%%%%%%%%%%%%%%%%%
\begin{tabular}{c}
\includegraphics[width=.3\linewidth, height=.3\linewidth]{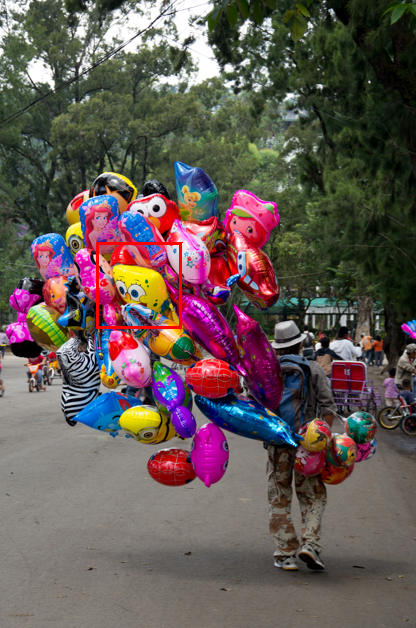}\\
 \scriptsize 022 from PIRM-self\\
\end{tabular}& 
\begin{tabular}{c}
\begin{tabular}{cccccc}
\includegraphics[width=.12\linewidth]{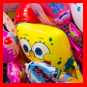}&
\includegraphics[width=.12\linewidth]{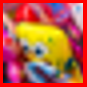}&
\includegraphics[width=.12\linewidth]{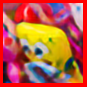}&
\includegraphics[width=.12\linewidth]{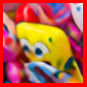}&
\includegraphics[width=.12\linewidth]{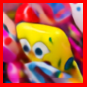}\\
\scriptsize $I_{\text{HR}}$ &\scriptsize Bicubic & \scriptsize SRCNN\cite{dong2016image} & \scriptsize  EDSR\cite{lim2017enhanced} &\scriptsize DBPN\cite{haris2018deep}\\
\includegraphics[width=.12\linewidth]{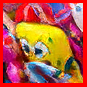}&
\includegraphics[width=.12\linewidth]{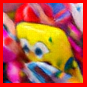}&
\includegraphics[width=.12\linewidth]{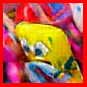}&
\includegraphics[width=.12\linewidth]{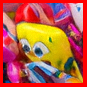}&
\includegraphics[width=.12\linewidth]{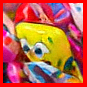}\\
 \scriptsize ENet\cite{sajjadi2017enhancenet} & \scriptsize BNet$_1$  & \scriptsize BNet$_3$ & \scriptsize  EPSR$_1$ & \scriptsize EPSR$_3$\\
\end{tabular}\\
\end{tabular}\\
%%%%%%%%%%IMAGE4%%%%%%%%%%%%%%%%%%%%%%%%
\begin{tabular}{c}
\includegraphics[width=.3\linewidth, height=.3\linewidth]{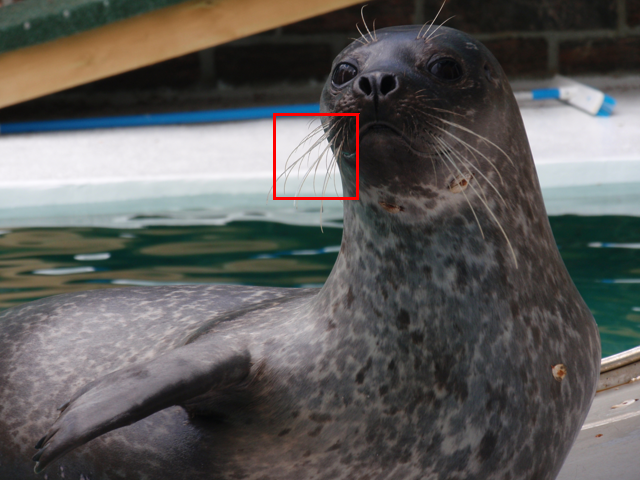}\\
\scriptsize 045 from PIRM-self\\
\end{tabular}&
\begin{tabular}{c}
\begin{tabular}{cccccc}
\includegraphics[width=.12\linewidth]{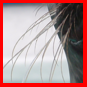}&
\includegraphics[width=.12\linewidth]{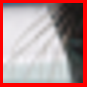}&
\includegraphics[width=.12\linewidth]{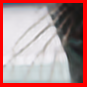}&
\includegraphics[width=.12\linewidth]{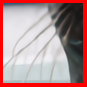}&
\includegraphics[width=.12\linewidth]{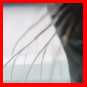}\\
\scriptsize $I_{\text{HR}}$ &\scriptsize Bicubic & \scriptsize SRCNN\cite{dong2016image} & \scriptsize  EDSR\cite{lim2017enhanced} &\scriptsize DBPN\cite{haris2018deep}\\
\includegraphics[width=.12\linewidth]{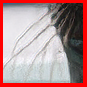}&
\includegraphics[width=.12\linewidth]{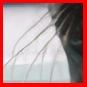}&
\includegraphics[width=.12\linewidth]{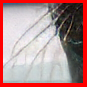}&
\includegraphics[width=.12\linewidth]{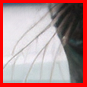}&
\includegraphics[width=.12\linewidth]{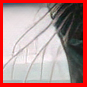}\\
\scriptsize ENet\cite{sajjadi2017enhancenet} & \scriptsize BNet$_1$ & \scriptsize BNet$_3$ & \scriptsize  EPSR$_1$ & \scriptsize EPSR$_3$\\
\end{tabular}\\
\end{tabular}\\ 
\end{tabular}\\
\caption{Qualitative comparison of our models with other works on x4 super-resolution. Examples are taken from datasets of Urban100 (Rows 1-2) and PIRM-self (Rows 3-4).}
\label{qual2}
\end{figure*}
\subsection{Qualitative Results}
For qualitative comparisons, we show a few examples from the standard benchmark datasets. In all the cases, we also show the ground truth (GT) images to get an idea about the content distortions introduced by the perceptual SR methods and also to visualize the extent to which the distortion measure based methods can reveal the lost details. Fig. \ref{qual1} and Fig. \ref{qual2} shows visual comparisons of seven examples in total. Examples in Fig. \ref{qual1} and Fig. \ref{qual2} clearly shows that, though ENet\cite{sajjadi2017enhancenet} is able to achieve a significant level of detail enhancement, the texture details added by the network is often very different from the ground-truth. Also, ENet\cite{sajjadi2017enhancenet} appears to add strong noise components while attempting to do detail enhancement. In comparison to ENet\cite{sajjadi2017enhancenet}, the presence of noise and unrealistic texture is less for the case of CX\cite{mechrez2018contextual} while maintaining a comparable level of detail enhancement. Contrarily, EPSR$_3$ is able to generate realistic textures that are faithful to both the GT image and the outputs from distortion-based methods.

The presence of spurious noise components in ENet\cite{sajjadi2017enhancenet} outputs can be seen in the first example of Fig. \ref{qual1} as well as the first and second example of Fig. \ref{qual2}. For all these examples, BNet$_3$ also resulted in a very similar noise disturbance. However, EPSR$_3$ was able to generate visually pleasing realistic textures in the output. Second and third examples in Fig. \ref{qual1} corresponds to failure case of ENet\cite{sajjadi2017enhancenet}, CX\cite{mechrez2018contextual}, and BNet$_3$ wherein all of them resulted in texture patterns that are very different from the GT, whereas EPSR$_3$ has succeeded in generating outputs that are more faithful to the GT image. The fourth example of Fig. \ref{qual2} shows the detail-preservation ability of EPSR as compared to the other perceptual methods. While EPSR$_3$ succeeded in reconstructing the seal whiskers to a great extent, both BNet and ENet\cite{sajjadi2017enhancenet} failed to do so.

In all the examples, the inadequacy of distortion based methods for reconstructing detailed textures is clearly evident. While outputs from both bicubic and SRCNN is affected by heavy blur, EDSR and DBPN output images with a minimal level of blur. The perceptual SR methods, on the other hand, generates detailed structures that are not necessarily consistent with the GT image. Among all the perceptual SR methods, EPSR performs the most convincing detail enhancement and is the one which generates detail enhanced outputs that are closest to the GT image. As indicated by the quantitative evaluation, EPSR$_1$ achieves significant perceptual quality improvement over EDSR while incurring only minimal distortion as compared to EDSR. This effect is predominantly visible in the first example of Fig. \ref{qual1} and first two examples from Fig. \ref{qual2}. As is evident from the Visual comparison of images from EDSR and EPSR reveals the progressive detail recovery that can be achieved by EPSR while moving across different regions in the perception-distortion plane. A very similar observation can also be made by comparing the images corresponding to BNet too. The source code and results of our method can be downloaded from \url{https://github.com/subeeshvasu/2018_subeesh_epsr_eccvw}.
\begin{figure*}
\centering
\begin{tabular}{cc}
\includegraphics[width=.455\linewidth,trim={0.0cm 0.0cm 0.0cm 0.0cm},clip]{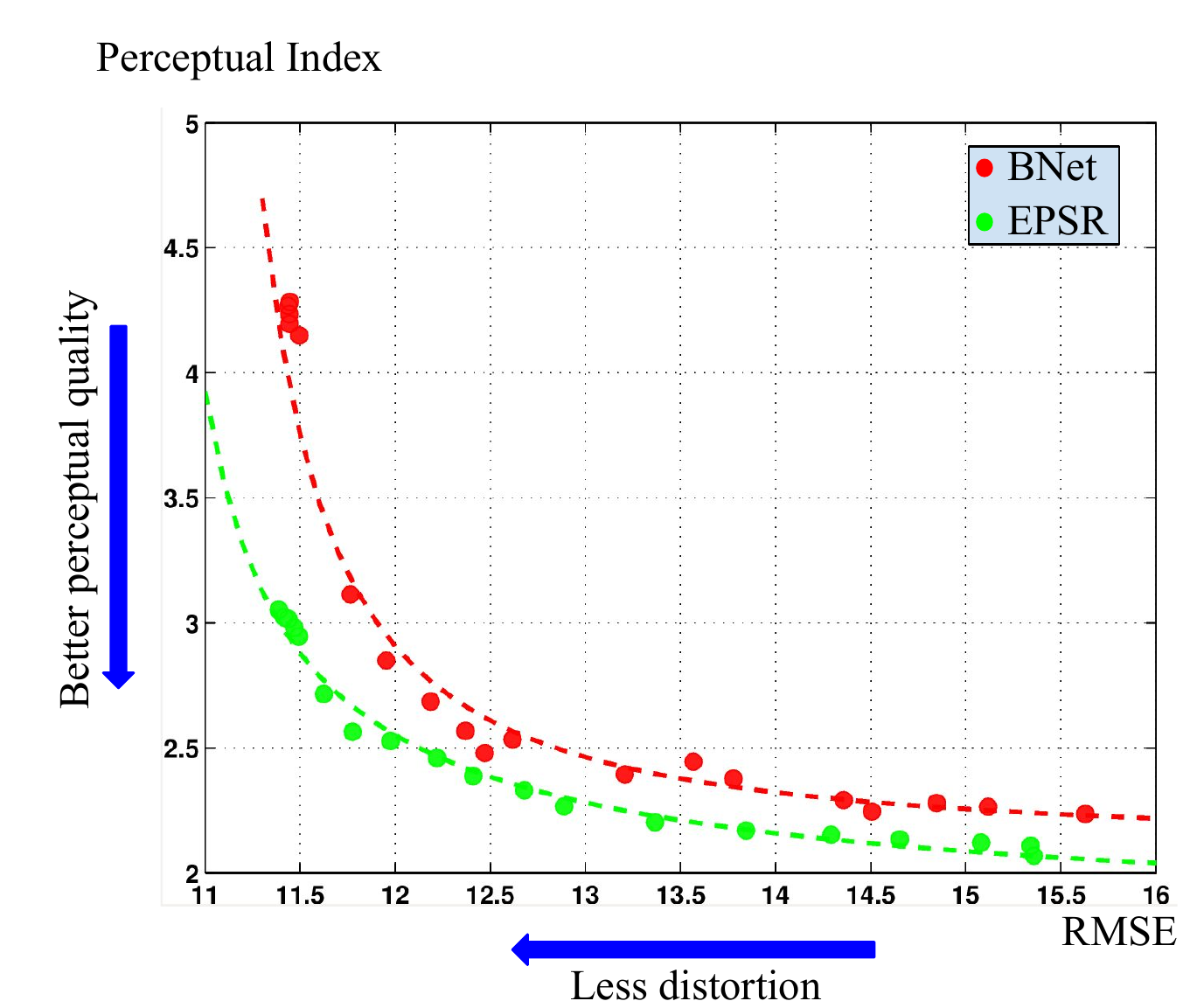}&
\includegraphics[width=.52\linewidth]{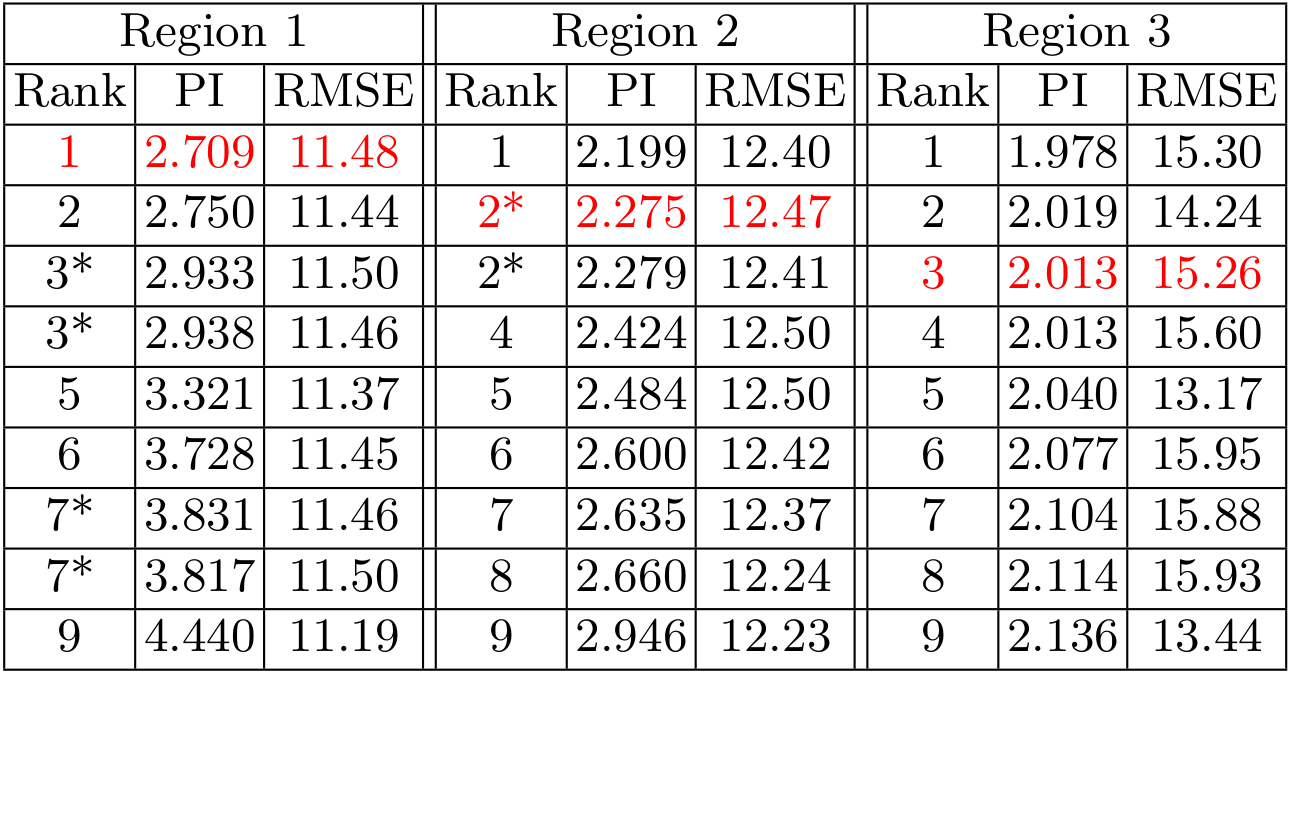}\\
(a)&(b)\\
\end{tabular}
\caption{(a)Perception-distortion trade-off between BNet and EPSR. For both methods, we plot the values corresponding to 19 model weights which span different regions on the perception-distortion plane and the corresponding curves that best fit these values. (b) Performance comparison of top 9 methods from PIRM-SR challenge \cite{2018arXiv180907517B}. Methods are ranked based on the PI and RMSE values corresponding to the test data of PIRM-SR. The entries from our approach are highlighted in red. Methods with a marginal difference in PI and RMSE values share the same rank and are indicated with a *.}
\label{trade-off-fig}
\end{figure*}
\subsection{Trade-off Comparison with BNet}
To analyze the impact of the generator module in achieving the trade-off, we perform a trade-off comparison between BNet and EPSR. Note that BNet uses a generator which is inferior to that of EPSR in terms of the distortion measures. Therefore, we expect to obtain a better perception-distortion trade-off using EPSR. Fig. \ref{trade-off-fig}(a) is a plot corresponding to the trade-off comparison between BNet and EPSR, wherein we have used the network model weights corresponding to different parameter settings that span different regions in the perception-distortion plane. To generate the plot in Fig. \ref{trade-off-fig}, we use the PI and RMSE values computed based on the PIRM-self dataset. To obtain model weights corresponding to different trade-off points, we have trained BNet and EPSR with different parameter settings and chose a number of network weights that yields the lowest PI values over a certain range of RMSE. It is evident from Fig. \ref{trade-off-fig} that EPSR is able to deliver a much better trade-off as compared to BNet as expected.
\subsection{PIRM challenge on perceptual super-resolution}
We have used our proposed model EPSR to participate in the PIRM-SR challenge \cite{2018arXiv180907517B} wherein the objective was to compare and rank perceptual SISR methods for an SR factor of 4. In order to rank each method, the perception-distortion plane was divided into three regions defined by thresholds on the RMSE. In each region, the winning algorithm is selected as the one that achieves the best perceptual quality. We have used parameter-tuned variants of EPSR to obtain the results corresponding to all three regions. The RMSE range used to define the three regions and the parameter settings that we have used to generate the results corresponding to the three regions are mentioned in Table \ref{region-table}. Our method was ranked 1,2, and 3 in region 1,2, and 3 respectively as shown in Fig. \ref{trade-off-fig}(b).
\section{Conclusions}
We proposed an extension to the state-of-the-art EDSR network by using it within a GAN framework. The proposed approach, EPSR, scales well in different regions of the perception-distortion plane and achieves superior perceptual scores when compared in a region-wise manner to other existing works. The performance improvement achieved by our approach is a cumulative result of the following factors: state-of-the-art SR network (EDSR) as the generator module, careful selection of loss function weights, and initialization of GAN training with the pretrained weights of EDSR. Our analysis of the perception-distortion trade-off between BNet and EPSR signal the possibility to further boost the trade-off by adopting another generator module that yields better distortion measures.
\clearpage

\bibliographystyle{splncs04}
\bibliography{epsr}
\end{document}